\documentclass[runningheads]{llncs}
\usepackage[T1]{fontenc}
\usepackage{graphicx}
\usepackage{array}
\usepackage{tabularx}
\usepackage{hyperref}
\usepackage{amsmath}
\usepackage{tikz}
\usepackage{algorithm}
\usepackage{algpseudocode}
\usepackage{booktabs}
\usepackage{progressbar}
\usepackage{makecell}

\usetikzlibrary{shapes.geometric, arrows.meta, positioning}

\usepackage{color}

\begin{document}
\title{PU classification under Non-SCAR: clustering-assisted logistic model with oversampling enhancement}
\titlerunning{PU classification under Non-SCAR ...}
\author{Konrad Furma\'{n}czyk\inst{1}\orcidID{0000-0002-7683-4787} \\
Kacper Paczutkowski\inst{1}\orcidID{0000-0001-7408-6060}}

\authorrunning{Furmańczyk K. et al.}
\institute{Institute of Information Technology, Warsaw University of Life Sciences, Warsaw, Poland
\email{\{konrad\_furmanczyk,kacper\_paczutkowski\}@sggw.edu.pl}}

\maketitle              

\begin{abstract}
This study addresses the PU classification problem under violations of the SCAR assumption. We investigate logistic regression–based approaches, namely the cluster method and its extensions with strict and non-strict Lasso regularization. The primary contribution of this work is the integration of the SMOTE technique to alleviate class imbalance and systematically assess its impact on the performance of the considered algorithms. SMOTE is first applied to rebalance the training dataset. Next, cleaning labels are derived via 2‑means clustering. Logistic regression is then trained on the cleaned data, where identified positive instances are augmented with additional true positives and the remaining observations are treated as negative. The experimental evaluation is conducted on 13 real benchmark datasets and one synthetic dataset. For comparison, we include the naive approach and the Spy-EM method. The results demonstrate that incorporating SMOTE improves classification performance when the SCAR condition is violated and indicate moderate robustness of the LassoJoint method in this setting.

\keywords{Positive-unlabeled learning  \and Logistic regression \and Clustering \and Oversampling}
\end{abstract}
\section{Introduction}
Positive–unlabeled (PU) learning constitutes a non-standard classification framework in which the training sample contains labeled positive examples and unlabeled observations only. The true class variable $Y \in \{0,1\}$ is not fully observed. Instead, one has access to a surrogate indicator $S \in \{0,1\}$ such that $S=1$ implies $Y=1$, while $S=0$ corresponds to unlabeled data. Consequently, negative labels are never explicitly observed, and the unlabeled subset is a mixture of positive and negative instances.

A typical example arises in medical diagnosis. Confirmed cases of a disease
correspond to \( S = 1 \) and thus \( Y = 1 \). However, individuals without
a confirmed diagnosis (\( S = 0 \)) may either be truly healthy (\( Y = 0 \))
or affected but undiagnosed (\( Y = 1 \)). This partial observability of \( Y \)
constitutes the fundamental difficulty in PU learning.
A similar situation occurs in fraud detection, where only confirmed fraudulent
transactions are labeled as positive, while the remaining transactions are
treated as unlabeled since they may include both legitimate and undiscovered
fraudulent cases. Further information on applications of PU learning can be found in \cite{Bek,C,L1,L2,L3,R,Yi}.

Formally, let $(Y_i,X_i,S_i)$, $i=1,\ldots,n$, be an i.i.d.\ sample from an unknown distribution $P(Y,X,S)$, where $X$ denotes a vector of covariates. Unlike standard supervised learning, only $(X_i,S_i)$ are observed. The PU mechanism assumes that labeling is restricted to positive examples, i.e. $P(S=1 \mid Y=0,X)=0,$ while for $Y=1$ the labeling probability may depend on additional assumptions about the selection process.

A widely studied scenario is the Selected Completely At Random (SCAR) assumption, $P(S=1 \mid Y=1,X)=P(S=1 \mid Y=1)=c,$ where $c$ is called the label frequency. Under SCAR, the labeling mechanism is independent of the features given the positive class, which leads to the simple relationship $P(Y=1 \mid X=x)=\frac{1}{c} P(S=1 \mid X=x).$
This property considerably simplifies estimation and explains the popularity of SCAR-based methods in the literature. In many real applications, however, the probability of labeling depends on the covariates. This more general setting is described by the Selected At Random (SAR) assumption, $P(S=1 \mid Y=1,X=x)=e(x),$ where $e(x)$ denotes the propensity score. In this case, the identification and estimation problems become substantially more challenging. 
A typical example where the SCAR assumption is violated arises in customer behavior analysis. Customers who leave explicit positive feedback or ratings correspond to labeled positives ($S=1$). However, the probability of providing feedback depends on customer characteristics such as engagement level, age, or purchasing frequency. As a result, positive examples are not selected completely at random, and highly active users are more likely to be labeled than less active ones. In such situations, the labeling mechanism depends on the features, and the SAR (Selected At Random) assumption is more appropriate than SCAR. Methodological developments under SAR include EM-based procedures \cite{Bek1,Gong} and joint maximum likelihood estimation for coupled logistic models of $e(x)$ and $P(Y=1 \mid X=x)$ \cite{F4}. Within the logistic regression paradigm, three main approaches have been proposed: the naive, weighted, and joint estimators \cite{Tes}. Extensions incorporating regularization, such as the LassoJoint method \cite{F1,F2,F3}, further enhance stability and variable selection under the SCAR framework. Although logistic regression remains attractive due to interpretability and computational efficiency, alternative strategies, including generative adversarial networks, have also been explored in recent PU research \cite{Hu,Gu}.
In this work, we investigate a computationally efficient cluster-based cleaning algorithm for the PU classification problem in the presence of violations of the SCAR assumption. Our approach builds upon logistic regression models combined with a clustering step used to identify reliable negative instances. In particular, we analyze the cluster method and its regularized extensions with strict and non-strict Lasso penalties.

A central component of the study is the incorporation of the SMOTE algorithm (\cite{Siriseriwan}) to mitigate class imbalance inherent in the PU framework. We systematically assess the impact of synthetic minority oversampling on the predictive performance and robustness of the cluster-based approaches. For benchmarking purposes, we additionally consider the naive estimator and the Spy-EM algorithm \cite{LX}. Furthermore, we investigate the robustness of the LassoJoint method under violations of the SCAR assumption, both with and without the application of SMOTE, assessing its stability and predictive performance in the more general SAR setting.

The remainder of our paper is structured as follows. In Section 2 we introduce our aims and methods, in Section 3 we present our numerical experiments together with the obtained results, in Section 4 we summarize and conclude our study. Comprehensive results of our study are available in Supplement \footnote{https://github.com/kapacc/mdai26}.
This supplement also includes all codes in R that are fully reproducible.
In order to execute our simulations, we employed the RStudio server module from the ICM UW Topola server\footnote{This research was carried out with the support of the Interdisciplinary Centre for Mathematical and Computational Modelling (ICM) at the University of Warsaw, under computational allocation No. g99-2175/g103-2563.}. We applied the following libraries:  glmnet \cite{Fr,glmnet},  caret \cite{Caret}, dplyr \cite{Dplyr}, naivebayes \cite{naivebayes}.

\section{Objectives and Methods}
This section defines the research objectives and details the methodological framework adopted in the study. First, we present the LassoJoint method, the naive approach, and Spy-EM. Next, we introduce our proposed cluster-based cleaning method, initially proposed in \cite{mdai}. Finally, we review the basic classification metrics that will be used for evaluation.

The primary goal of this work is to examine the performance of a computationally efficient cluster-cleaning algorithm when combined with SMOTE, and to assess its effectiveness in addressing the PU classification problem under violations of the SCAR assumption. The secondary objective is to evaluate the robustness of the LassoJoint method under such SCAR perturbations.
\\The \textit{LassoJoint} method \cite{F1} can be described as follows:
(1) For available PU dataset $(S_{i},X_{i})$, $i=1,\ldots ,n$, we perform
the ordinary Lasso procedure (see Tibshirani \cite{T}) for some tuning
parameter $\lambda >0$, i.e. we compute the following Lasso estimator of $%
\beta ^{\ast }=-argmin_{\beta}E_{X,Y}l(\beta,X,Y)$ and $l$ is the logistic loss (see \cite{F3} for more details)
$$\widehat{\beta}^{(L)}=\arg \min_{\beta\in R^{p+1}}\widehat{R}(\beta )+\lambda
\sum_{j=1}^{p}\left\vert \beta _{j}\right\vert,$$
where $$\widehat{R}(\beta )=-\frac{1}{n}\sum_{i=1}^{n}\left[ S_{i}\log
\left( \sigma (X_{i}^{T}\beta )\right) +\left( 1-S_{i}\right) \log \left(
1-\sigma (X_{i}^{T}\beta )\right) \right],$$ where $\sigma(t)=exp(t)/(1+exp(t))$
and subsequently, we obtain the corresponding support $Supp^{(L)}=\{1\leq
j\leq p:\widehat{\beta}_{j}^{(L)}\neq 0\}$;
(2) We perform the thresholded Lasso for some prespecified level $\delta $
and obtain the support $Supp^{(TL)}=\{1\leq j\leq p:\left\vert \widehat{\beta}%
_{j}^{(L)}\right\vert \geq \delta \}$;
(3) We apply the joint method from Teisseyre et al. \cite{Tes} for the
predictors from $Supp^{(TL)}$.
%\newline
The joint method minimizes, with respect to the parameter vector $b$
and the label frequency $c$, the empirical risk
\[
\widehat{R}(b,c)
= -\frac{1}{n}\sum_{i=1}^{n}
\Big[
S_i \log\!\big(c\sigma(X_i^Tb)\big)
+ (1-S_i)\log\!\big(1-c\sigma(X_i^Tb)\big)
\Big].
\]
The estimator of the posterior probability is 
\[
\widehat{f}_{\text{joint}}(x)
= \widehat{c}_{\text{joint}}^{-1}
\sigma(x^T\widehat{b}_{\text{joint}}),
\]
with $(\widehat{b}_{\text{joint}},\widehat{c}_{\text{joint}})
=\arg\min_{b,c}\widehat{R}(b,c).$
It should be emphasized that, under mild regularity conditions and SCAR assumption, the LassoJoint procedure satisfies the screening property; that is, with high probability, all significant predictors are selected in the first two steps (see Theorem~1(b) in \cite{F1}).
This method includes a selection step and can be applied to both high-dimensional and low-dimensional data.
\\The \textit{naive method} is when we applied logistic regression for $S=1$ and $S=0$ naive considering surrogate labels $S$ as true labels for $Y$.
\newline
We employ also the two‑step \textit{Spy‑EM} algorithm, one of the classical and most influential methods in the field of positive–unlabeled (PU) learning. This approach was originally introduced in the seminal work \cite{spyEM} on partially supervised text classification, which laid the foundations for many subsequent PU methodologies. The algorithm combines the spy technique with an iterative EM procedure, enabling robust estimation of the distributions of positive and unlabeled examples. In our study, we use Spy‑EM as a strong baseline representing established, well‑grounded PU learning techniques.

In addition to mitigate of partially labeled target, synthetic oversampling was applied to enrich the positive class and reduce the distortion caused by missing labels. We chose the SMOTE (Synthetic Minority Over-sampling Technique) algorithm from a relatively recent implementation \cite{Siriseriwan}. In out experiments, we employed the default configuration with K=5 and dup\_size = 0, which corresponds to using five nearest neighbours during synthetic sample generation and enabling the automatic oversampling mode that balances the minority class to a level comparable with the majority class.

\subsection{Cluster Cleaning Procedure}
In \cite{mdai}, we proposed a cluster-based cleaning method, which we call "pecking" due to its iterative extraction of subsets. Briefly, a fraction $q\%$ of labeled positives ($S=1$) is temporarily added to the unlabeled set ($S=0$), forming $C_0^q$. Two-means clustering is applied to $C_0^q$, producing clusters $C_{0,0}$ and $C_{0,1}$, with $C_{0,1}$ containing more $S=1$ labels. Labels are assigned as $\hat{Y}=1$ for $\{S=1\} \cup C_{0,1}$ and $\hat{Y}=0$ for $C_{0,0}$. The procedure is repeated $R$ times to reduce variance and improve stability.

For each repetition, a logistic regression model is fitted on $(\hat{Y}_i,X_i)$. In addition, Lasso regularization is applied in two variants: \textit{strict} (averaging coefficients of features appearing in all repetitions) and \textit{non-strict} (averaging coefficients of features appearing in any repetition). The final predictive model is obtained by averaging coefficients across repetitions, effectively aggregating the results of multiple "pecked" subsets. To mitigate class imbalance inherent in the PU setting, the SMOTE algorithm is incorporated during the preprocessing of the train data, which synthetically generates additional positive instances and improves the stability and predictive performance of the cluster-based approaches. 
The pseudocode for this algorithm is presented below.

\begin{algorithm}[h]
\caption{Pecking Procedure for PU Learning}
\begin{algorithmic}[1]
\Require Dataset $(X_i, S_i)$ for $i = 1, \ldots, n$ with $S_i \in \{0,1\}$, proportion $q \in (0,1)$, repetitions $R$
\Ensure Final predictive model coefficients

\For{$r = 1$ to $R$}
    \State Draw $q\%$ of samples with $S=1$ and add to samples with $S=0$ to form $C_0^q$
    \State Apply 2-means clustering on $C_0^q$ to get clusters $C_{0,0}$ and $C_{0,1}$

    \If{$C_{0,0}$ has more samples with $S=1$ than $S=0$}
        \State Swap $C_{0,0} \leftrightarrow C_{0,1}$ \Comment{Ensure $C_{0,1}$ has more $S=1$}
    \EndIf

    \State Assign labels:
    \State \quad $\hat{Y} = 1$ for all samples in $\{S=1\} \cup C_{0,1}$
    \State \quad $\hat{Y} = 0$ for all samples in $C_{0,0} \setminus \{S=1\}$

    \State Fit either:
    \State \quad Logistic regression on $(X_i, \hat{Y}_i)$
    \State \quad or Lasso (strict / non-strict variant)

    \State Store resulting coefficients $\beta^{(r)}$
\EndFor

\State Compute final coefficients:
\If{Logistic regression}
    \State Average all $\beta_j$ across $R$ repetitions
\ElsIf{Lasso (strict)}
    \State Average $\beta_j$ only for features appearing in \textbf{all} $R$ repetitions
\Else \Comment{Lasso non-strict}
    \State Average $\beta_j$ for features appearing in \textbf{any} repetition
\EndIf

\end{algorithmic}
\end{algorithm}

To date, the PU research literature has employed a few approaches based on clustering methods. For instance, in \cite{LX},
in the context of text classification, the initial step of this approach involves the collection of reliable negative examples (C-CRNE) through hierarchical clustering. 
During the process of building the classifier, the Term Frequency Inverse Positive-Negative Document Frequency approach is adopted.
In contrast to the aforementioned approach, the proposed method does not exclusively focus on text classification. 
It is also computationally more efficient and expeditious due to its utilisation of the 2-means method in place of hierarchical clustering.

%\subsection{Classification Metrics}
%The accuracy and F1-score are defined as follows:
%$$Accuracy=\frac{TP+TN}{TP+FP+FN+TN},$$
%$$Recall = \frac{TP}{TP + FN}, $$
%$$Precision = \frac{TP}{TP + FP},$$  
%$$F1 = \frac{2 \cdot Precision \cdot Recall }{Precision + Recall},$$ 
%where $TP,$ $FN,$ $TN$ and $FP$ stand for: the number of true positives, false negatives, true negatives and false positives, respectively. We will also report the AUC value, a summary metric of the ROC curve that reflects the test's ability to distinguish between positive and negative individuals.
\section{Numerical Experiments}
\subsection{Datasets}
The metrics for the naive method and the cluster-based method (in two scenarios) were evaluated on one synthetic dataset and 13 real low-dimensional datasets from the UCI Machine Learning Repository \cite{uci}. Our synthetic dataset (Artif) was generated by creating a matrix $X$ with both relevant and irrelevant features, where the first variables had increasing standard deviations. A logistic function was applied to the relevant features to probabilistically generate the binary target $Y$.

We removed quasi-constant features (most common value >90\%) and highly correlated features to reduce redundancy. Numeric features were Min-Max scaled, and only variables with at least 5 unique values were retained to preserve continuous features. The summary table (Table 1 in Supplementary Materials\footnote{\url{https://github.com/kapacc/mdai26/blob/main/supplement.pdf}}) outlines key dataset characteristics—such as feature counts, class distribution, and mean absolute correlation—showing that the datasets differ in size and complexity, with variables having fewer than 15 unique values treated as non‑continuous.

\subsection{Non-SCAR Scenarios}
In the SCAR (Selection Completely At Random) scheme, the coefficient $c$ can be set in advance, since the selection is independent of the features. In the non-SCAR scheme, generating $S$ at a desired $c$ requires a specific procedure, as selection may depend on feature values. In both cases, $c$ can be computed a posteriori from the empirical distribution, but only in SCAR can it be fixed a priori. For our non-SCAR experiments, we generate the surrogate variable $S$ using high-variance features. The top $n_{\text{vars}}$ columns are selected, and intermediate variables based on cumulative sums of $Y$ are computed. A linear approximation allows us to control the coverage of $Y$ by $S$, producing a dataset with a value close to the target $c$ (see Supplementary Materials, Section 4, \texttt{non\_scar\_labelling.mvc}). In our experiments, we set $n_{\text{vars}} = 2$. This method enables control over problem difficulty while modeling a complex dependence between the surrogate and the true explanatory variable.

\subsection{Calibrations of Parameters and Experiment}
We employed the naive logistic regression, Spy-EM, Pecking procedure: cluster method, and LassoJoint (strict and non-strict) to address PU classification under the non-SCAR assumption. For each fully labeled dataset, we randomly select $c\cdot 100\%$ of labeled observations ($S=1$) for $c = 0.3, 0.5, 0.8$, and split the data into training ($70\%$) and test ($30\%$) sets. For the cluster cleaning method, GLM coefficients are averaged over $R=5$ repetitions. Following \cite{mdai}, we set $q=1$, which produced the most consistent results. For completeness we also provide results for $q=0.25,0.5,1$ in the supplementary material, Sections 1.1 and 2.1. Each method is evaluated with and without SMOTE. Spy‑EM is used however in its original form.

For LassoJoint, the first step uses Lasso with tuning parameter $\lambda$ chosen via 10-fold cross-validation (using $\lambda_{\min}$). In the second step, thresholded Lasso is applied with $\delta = 0.5 \cdot \lambda$ based on the first step \cite{F3}. Classification metrics are then computed from Monte Carlo replications. The number of repetitions depends on dataset size: up to 100 for datasets with fewer than 100 features or 10,000 rows, and 10 repetitions for larger datasets. This balances thoroughness on small datasets with efficiency on large ones.
\begin{figure}[!ht]
    \centering
    \includegraphics[width=\textwidth]{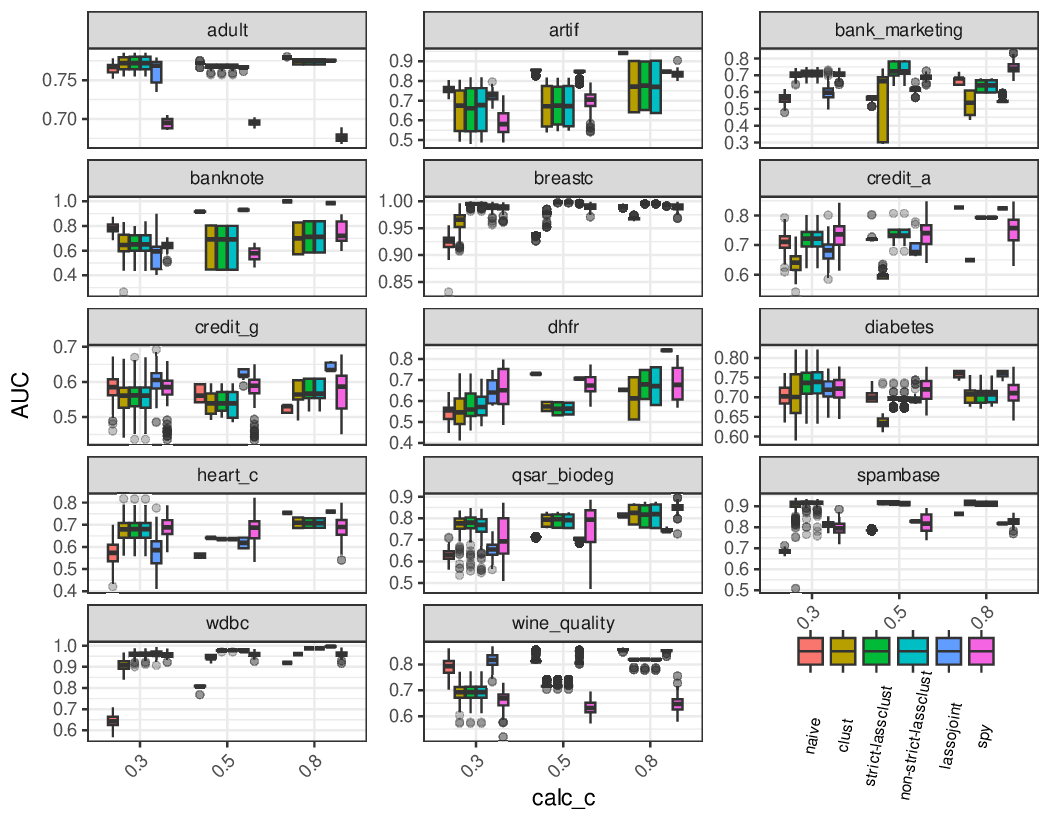}
    \caption{AUC; SMOTE scheme; nonSCAR}
\end{figure}
\subsection{Results}
In the present simulation study, both non-SCAR and SCAR assumptions were considered.  The SMOTE method improved the AUC in 69\% of the experiments for the clust method and in approximately 56\% of the experiments for the Lasso variants (strict and non-strict). In contrast, the improvement was least pronounced for the naive method, where AUC increased in only about 16\% of the cases. With respect to Accuracy, SMOTE improved the results in 72\% of the experiments for the clust method, in approximately 67\% for the Lasso variants (strict and non-strict), and to a lesser extent for the LassoJoint method (27\%; see Table~2). Table~3 reports the percentage differences between the SMOTE and non-SMOTE versions of the considered methods under both SCAR and non-SCAR sampling schemes. Several conclusions can be drawn from these results. First, SMOTE provides substantial improvements for clustering-based approaches. In particular, the methods \textit{clust}, \textit{strict-lassclust}, and \textit{non-strict-lassclust} consistently achieve noticeable gains in AUC across all settings, with improvements increasing as the parameter $c$ grows. The largest gains are observed for $c=0.8$, where AUC improvements exceed $10\%$ and accuracy improvements reach up to $60\%$. Second, the benefits of SMOTE are significantly stronger for classification accuracy than for AUC. While AUC improvements are moderate but stable, accuracy gains can be very large, especially under the SCAR setting. This suggests that SMOTE primarily improves class balance and decision boundaries rather than ranking performance alone. Third, the naive method benefits from SMOTE mainly in terms of accuracy under SCAR, but shows negligible or even negative changes in AUC, indicating limited robustness of simple approaches to synthetic oversampling. Finally, the \textit{lassojoint} method does not benefit from SMOTE and in several cases exhibits slight performance degradation. This indicates that joint Lasso regularization is relatively insensitive to class imbalance or may already incorporate implicit regularization effects that reduce the usefulness of synthetic samples. In some cases, SMOTE further enhanced all metrics for cluster-based methods, which then outperformed naive and Spy-EM approaches. The LassoJoint method remained highly stable.

\begin{table}[h]
\caption{Number of experiments by SMOTE effect (AUC and Accuracy)}
\centering
\resizebox{0.75\textwidth}{!}{%
\begin{tabularx}{\textwidth}{lXXX}
\toprule
Method & Improved & Worsened & Equal\\
\midrule
\addlinespace[0.3em]
\multicolumn{4}{l}{\textbf{AUC}}\\
\hspace{1em}naive & \makecell[c]{\progressbar{0.164000} \\ 16.4\% (600)} & \makecell[c]{\progressbar{0.323000} \\ 32.3\% (1183)} & \makecell[c]{\progressbar{0.513000} \\ 51.3\% (1877)}\\
\hspace{1em}clust & \makecell[c]{\progressbar{0.690000} \\ 69\% (2526)} & \makecell[c]{\progressbar{0.218000} \\ 21.8\% (798)} & \makecell[c]{\progressbar{0.092000} \\ 9.2\% (336)}\\
\hspace{1em}strict-lassclust & \makecell[c]{\progressbar{0.540000} \\ 54\% (1976)} & \makecell[c]{\progressbar{0.292000} \\ 29.2\% (1067)} & \makecell[c]{\progressbar{0.169000} \\ 16.9\% (617)}\\
\hspace{1em}non-strict-lassclust & \makecell[c]{\progressbar{0.566000} \\ 56.6\% (2073)} & \makecell[c]{\progressbar{0.289000} \\ 28.9\% (1059)} & \makecell[c]{\progressbar{0.144000} \\ 14.4\% (528)}\\
\hspace{1em}lassojoint & \makecell[c]{\progressbar{0.354000} \\ 35.4\% (1295)} & \makecell[c]{\progressbar{0.327000} \\ 32.7\% (1195)} & \makecell[c]{\progressbar{0.320000} \\ 32\% (1170)}\\
\addlinespace[0.3em]
\multicolumn{4}{l}{\textbf{Accuracy}}\\
\hspace{1em}naive & \makecell[c]{\progressbar{0.400000} \\ 40\% (1464)} & \makecell[c]{\progressbar{0.120000} \\ 12\% (440)} & \makecell[c]{\progressbar{0.480000} \\ 48\% (1756)}\\
\hspace{1em}clust & \makecell[c]{\progressbar{0.719000} \\ 71.9\% (2630)} & \makecell[c]{\progressbar{0.148000} \\ 14.8\% (543)} & \makecell[c]{\progressbar{0.133000} \\ 13.3\% (487)}\\
\hspace{1em}strict-lassclust & \makecell[c]{\progressbar{0.679000} \\ 67.9\% (2484)} & \makecell[c]{\progressbar{0.208000} \\ 20.8\% (760)} & \makecell[c]{\progressbar{0.114000} \\ 11.4\% (416)}\\
\hspace{1em}non-strict-lassclust & \makecell[c]{\progressbar{0.669000} \\ 66.9\% (2450)} & \makecell[c]{\progressbar{0.205000} \\ 20.5\% (752)} & \makecell[c]{\progressbar{0.125000} \\ 12.5\% (458)}\\
\hspace{1em}lassojoint & \makecell[c]{\progressbar{0.274000} \\ 27.4\% (1004)} & \makecell[c]{\progressbar{0.309000} \\ 30.9\% (1131)} & \makecell[c]{\progressbar{0.417000} \\ 41.7\% (1525)}\\
\bottomrule
\end{tabularx}
}
\end{table}

\begin{table}[ht]
\centering
\caption{SMOTE vs non-SMOTE — \% diff. for SCAR and nonSCAR scheme}
\begin{tabular}{lrrrr}
\hline
Method & $\Delta$AUC (scar) & $\Delta$AUC (nonscar) & $\Delta$ACC (scar) & $\Delta$ACC (nonscar) \\
\hline
\multicolumn{5}{l}{\textbf{calc\_c = 0.3}} \\
naive & 0.1\% & -1.3\% & 82.4\% & 10.3\% \\
clust & 7.7\% & 2.1\% & 16.3\% & 7.3\% \\
strict-lassclust & 7.5\% & 1.2\% & 15.8\% & 6.1\% \\
non-strict-lassclust & 7.4\% & 1.4\% & 16.3\% & 7.8\% \\
lassojoint & -0.5\% & -0.3\% & 2.6\% & 9.1\% \\
\hline
\multicolumn{5}{l}{\textbf{calc\_c = 0.5}} \\
naive & -0.1\% & -0.3\% & 24.8\% & 4.4\% \\
clust & 7.2\% & 2.4\% & 23.5\% & 15.3\% \\
strict-lassclust & 7.5\% & -0.8\% & 21.8\% & 14.4\% \\
non-strict-lassclust & 8.1\% & -0.8\% & 23.9\% & 11.2\% \\
lassojoint & -0.4\% & -0.1\% & -2.8\% & -2.2\% \\
\hline
\multicolumn{5}{l}{\textbf{calc\_c = 0.8}} \\
naive & 0.0\% & 0.4\% & -2.3\% & -0.7\% \\
clust & 12.7\% & 10.6\% & 54.8\% & 30.0\% \\
strict-lassclust & 11.5\% & 7.4\% & 58.9\% & 28.3\% \\
non-strict-lassclust & 11.6\% & 7.7\% & 60.8\% & 27.9\% \\
lassojoint & 0.0\% & 0.1\% & 4.0\% & -0.7\% \\
\hline
\end{tabular}
\end{table}

%Under SCAR, the best metrics were obtained by LassoJoint, as expected since it was designed for this scenario. The naive and cluster methods also performed well, except for the Artif, Banknotes, and DHLR datasets (see Supplementary Materials, Section 2). Similar to the non-SCAR case with $n_{\text{vars}}=1$ in \cite{mdai}, using $n_{\text{vars}}=2$ yielded comparable results.

Clustering-based methods combined with regularization (strict and non-strict Lasso) show strong performance under the non-SCAR assumption and outperform the naive and Spy-EM approaches (see Fig.~1). The results obtained under the non-SCAR setting lead to several important conclusions. First, methods combining clustering with regularization (strict and non-strict Lasso variants) generally achieve competitive or superior performance compared to the naive approach, particularly in terms of Accuracy and F1-score. Second, the naive method tends to remain competitive in terms of AUC, especially for larger values of the labeling frequency parameter $c$. This behavior is expected, since increasing $c$ makes the PU learning problem closer to a fully labeled classification task, reducing the impact of selection bias. Third, the LassoJoint method exhibits stable performance across different datasets but does not consistently outperform clustering-based approaches in the non-SCAR scenario, suggesting that methods designed under the SCAR assumption may lose their advantage when this assumption is not satisfied. Moreover, the Spy-EM method rarely achieves the best results and is often dominated by clustering-based or regularized approaches across all considered performance measures. Finally, the application of SMOTE frequently improves classification performance, especially for clustering-based and Lasso-based methods, with noticeable gains in Accuracy and F1-score. The improvement is less pronounced for the naive method, indicating that imbalance handling is particularly beneficial when combined with structure-aware PU learning procedures. Overall, the results indicate that under non-SCAR conditions, hybrid approaches integrating clustering and regularization provide the most robust and reliable performance across diverse datasets.

\begin{table}[h!]
\centering
\caption{\% diff. methods (with SMOTE enhancement) vs naive in SCAR and nonSCAR scheme}
\label{tab:improvement_naive}
\begin{tabular}{lrrrr}
\hline
Method & $\Delta$AUC (scar) & $\Delta$AUC (nonscar) & $\Delta$ACC (scar) & $\Delta$ACC (nonscar) \\
\hline
\multicolumn{5}{l}{\textbf{calc\_c = 0.3}} \\
clust & -5.9\% & 7.4\% & 15.2\% & 30.5\% \\
lassojoint & -0.3\% & 5.1\% & - & -16.0\% \\
non-strict-lassclust & -5.0\% & 9.6\% & 14.8\% & 30.0\% \\
strict-lassclust & -5.0\% & 9.7\% & 15.0\% & 30.3\% \\
\hline
\multicolumn{5}{l}{\textbf{calc\_c = 0.5}} \\
clust & -7.6\% & -2.6\% & 15.5\% & 14.8\% \\
lassojoint & 0.3\% & 4.2\% & 15.3\% & -17.8\% \\
non-strict-lassclust & -6.3\% & 0.8\% & 14.2\% & 12.4\% \\
strict-lassclust & -6.2\% & 0.8\% & 14.5\% & 16.2\% \\
\hline
\multicolumn{5}{l}{\textbf{calc\_c = 0.8}} \\
clust & -4.0\% & -3.6\% & -1.5\% & 2.2\% \\
lassojoint & 0.3\% & -1.4\% & -3.7\% & -14.9\% \\
non-strict-lassclust & -3.6\% & -1.6\% & -1.6\% & 1.3\% \\
strict-lassclust & -3.6\% & -1.6\% & -1.7\% & 2.2\% \\
\hline
\end{tabular}
\end{table}

The results obtained under the SCAR assumption lead to several consistent observations. First, the LassoJoint method, which is explicitly designed under the SCAR framework, typically achieves the most stable and often the best overall performance across the considered datasets, particularly in terms of Accuracy and F1-score. This confirms the theoretical advantage of jointly estimating model parameters and the labeling frequency when the SCAR condition holds. Second, the naive approach remains highly competitive in terms of AUC, especially for larger values of the labeling frequency parameter $c$. As $c$ increases, the PU learning problem approaches a standard supervised classification setting, which naturally favors simpler estimators. Third, clustering-based methods combined with Lasso regularization provide competitive results but generally do not outperform LassoJoint under SCAR conditions, indicating that their main advantage appears when the SCAR assumption is violated. Finally, the incorporation of SMOTE frequently improves Accuracy and F1-score, particularly for clustering-based procedures, while its impact on AUC is usually moderate.

Full results for all metrics are provided in the Supplementary Materials (Section 1, Non-SCAR Scheme and Section 2, SCAR Scheme).
 \footnote{\url{https://github.com/kapacc/mdai26/blob/main/supplement.pdf}}. 
% \begin{figure}[h]
%     \centering
%     \includegraphics[width=\textwidth]{boxplots_time_values.eps}
%     \caption{Boxplots of Executing time for the methods}
%     \label{fig:boxplots}
% \end{figure}
\section{Conclusions}
In the considered model of SCAR disturbance, clustering methods combined with Lasso were effective for PU classification, and Lasso regularization generally improved predictions. Almost all methods, especially when combined with SMOTE, showed improved classifier performance, with some exceptions for LassoJoint. As $c$ increases, the PU problem approaches a fully labeled scenario, so for $c$ close to 1, the naive method gradually gains an advantage. As shown in Table~\ref{tab:improvement_naive} our methods deliver the strongest improvements for small values of c, where the gains in both AUC and ACC are most pronounced. As c increases, the advantage naturally diminishes, since the problem becomes closer to standard modeling without additional structure. It is also evident that the improvements tend to be slightly larger for the nonscar cases, indicating that our approach is particularly effective in that subset. Under SCAR, the LassoJoint algorithm, designed for this condition, performed nearly optimally. The cluster method also performed well, suggesting its applicability regardless of the SCAR assumption. This work extends the results from \cite{F2}--\cite{F3} to settings where SCAR is not satisfied. Finally, the first step of our procedure, based on 2-means clustering, can be applied not only to single-sample PU scenarios but also to case-control PU learning\cite{Bek}.
\subsubsection{\discintname}
The authors declare that they have no known competing financial interests or personal relationships that could have appeared to
influence the work reported in this paper.
%\end{credits}

\end{document}